\documentclass[10pt,twocolumn,letterpaper]{article}

\usepackage[pagenumbers]{cvpr} % To force page numbers, e.g. for an arXiv version
\usepackage{amsthm}
\usepackage{algorithm}
\usepackage{algorithmic}
\usepackage{times}
\usepackage{helvet}
\usepackage{courier}
\usepackage{xcolor}
\usepackage{amsfonts, amsmath}
\usepackage{multirow}
\usepackage{booktabs}
\newtheorem{theorem}{Theorem}

\theoremstyle{definition}

\usepackage{algorithm}
\usepackage{algorithmic}
\usepackage{multirow}
\usepackage[table]{xcolor}

\definecolor{cvprblue}{rgb}{0.21,0.49,0.74}
\usepackage[pagebackref,breaklinks,colorlinks,allcolors=cvprblue]{hyperref}

\def\paperID{12126} % *** Enter the Paper ID here
\def\confName{CVPR}
\def\confYear{2026}

\title{Towards Efficient VLMs: Information-Theoretic Driven Compression via Adaptive Structural Pruning}

\author{
    Zhaoqi Xu\textsuperscript{1}\\
    {\tt\small xuzhaoqi@mail.bnu.edu.cn}  % 1号作者邮箱
    \and
    Yingying Zhang\textsuperscript{2}\\
    {\tt\small yingyingzhang239@163.com}
    \and
    Jian Li\textsuperscript{1}\thanks{Corresponding author}\\
    {\tt\small jli@bnu.edu.cn}
    \and
    Jianwei Guo\textsuperscript{1}\\
    {\tt\small jianwei.guo@bnu.edu.cn}
    \and
    Qiannan Zhu\textsuperscript{1}\\
    {\tt\small zhuqiannan@bnu.edu.cn} 
    \and 
    Hua Huang\textsuperscript{1}\\
    {\tt\small huahuang@bnu.edu.cn} 
    \\ 
    \textsuperscript{1}School of Artificial Intelligence, Beijing Normal University, 
    \and
    \textsuperscript{2}Zhongtai Securities Institute for Financial Studies, Shandong University
}

\begin{document}
\maketitle
\begin{abstract}
Recent advances in vision-language models (VLMs) have shown remarkable performance across multimodal tasks, yet their ever-growing scale poses severe challenges for deployment and efficiency. Existing compression methods often rely on heuristic importance metrics or empirical pruning rules, lacking theoretical guarantees about information preservation.
In this work, we propose InfoPrune, an information-theoretic framework for adaptive structural compression of VLMs. Grounded in the Information Bottleneck principle, we formulate pruning as a trade-off between retaining task-relevant semantics and discarding redundant dependencies. To quantify the contribution of each attention head, we introduce an entropy-based effective rank (eRank) and employ the Kolmogorov–Smirnov (KS) distance to measure the divergence between original and compressed structures. This yields a unified criterion that jointly considers structural sparsity and informational efficiency.
Building on this foundation, we further design two complementary schemes: (1) a training-based head pruning guided by the proposed information loss objective, and (2) a training-free FFN compression via adaptive low-rank approximation.
Extensive experiments on VQAv2, TextVQA, and GQA demonstrate that InfoPrune achieves up to 3.2× FLOP reduction and 1.8× acceleration with negligible performance degradation, establishing a theoretically grounded and practically effective step toward efficient multimodal large models.
\end{abstract}    

\section{Introduction}
\label{sec:intro}

Vision-Language Models (VLMs)~\cite{wang2023visionllm, li2023evaluating, wang2024qwen2, cui2024survey, bai2025qwen2, meta2025llama} leverage the Transformer architecture~\cite{vaswani2017attention} to align visual and textual representations, achieving remarkable performance on tasks such as image captioning, visual question answering, and cross-modal retrieval~\cite{lu2019vilbert, radford2021learning, li2022blip}. However, as model scales continue to expand, VLMs face mounting practical challenges including computational redundancy, inference latency, and deployment overhead~\cite{bordes2024introduction}. These issues are particularly evident within the visual modules, whose design inefficiencies often limit both scalability and runtime efficiency~\cite{li2025survey, zhu2024survey}.

\begin{figure}[t]
    \centering
    \includegraphics[width=1.0\linewidth]{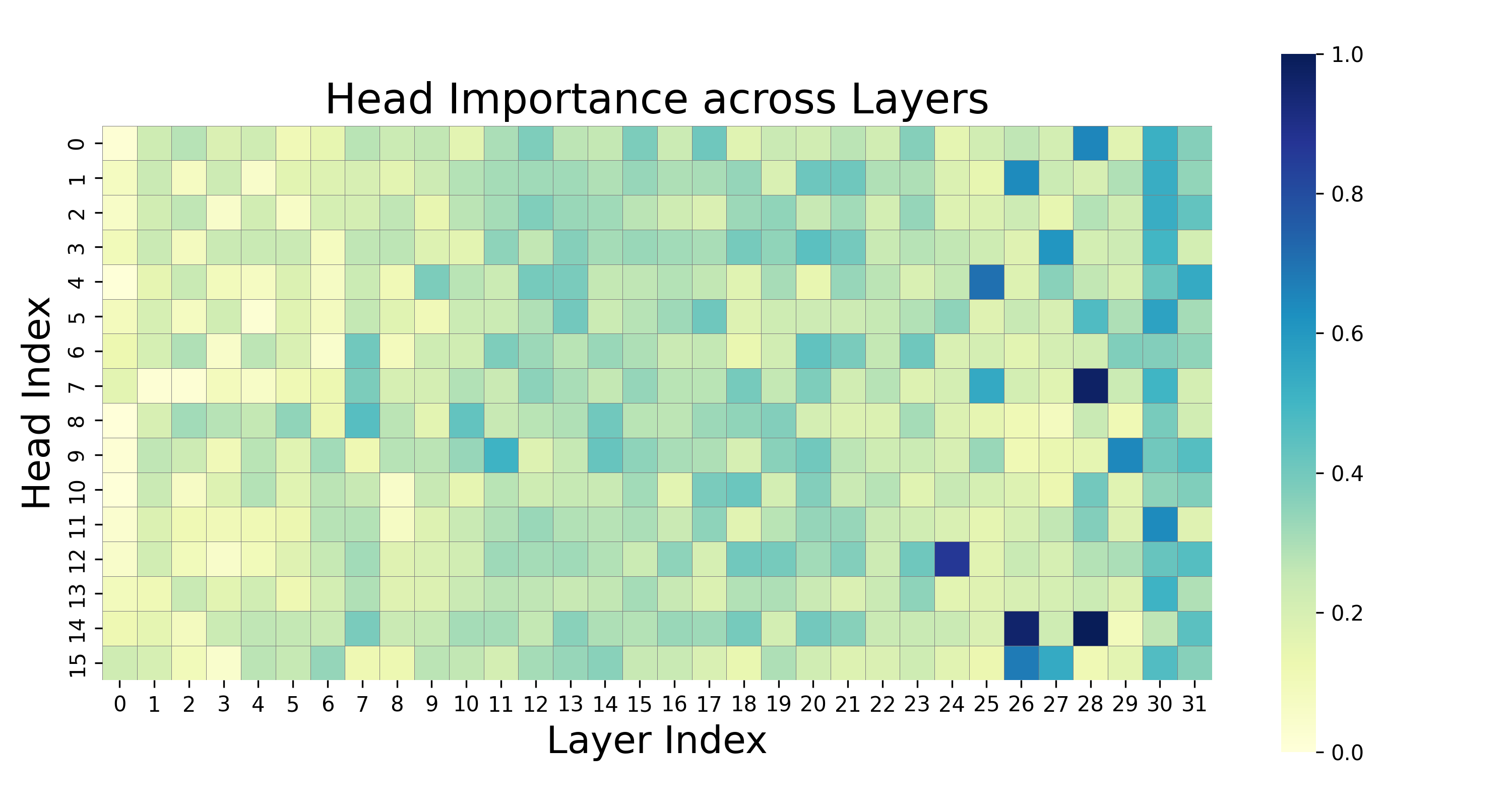}
    \caption{Normalized importance scores of attention heads across 32 layers in the Qwen2VL-7B visual modules. The scores are derived via min--max normalization of negative entropy values, reflecting each head’s relative contribution.}
    \label{attention_importance}
\end{figure}

\begin{figure}[t]
    \centering
    \includegraphics[width=1.0\linewidth]{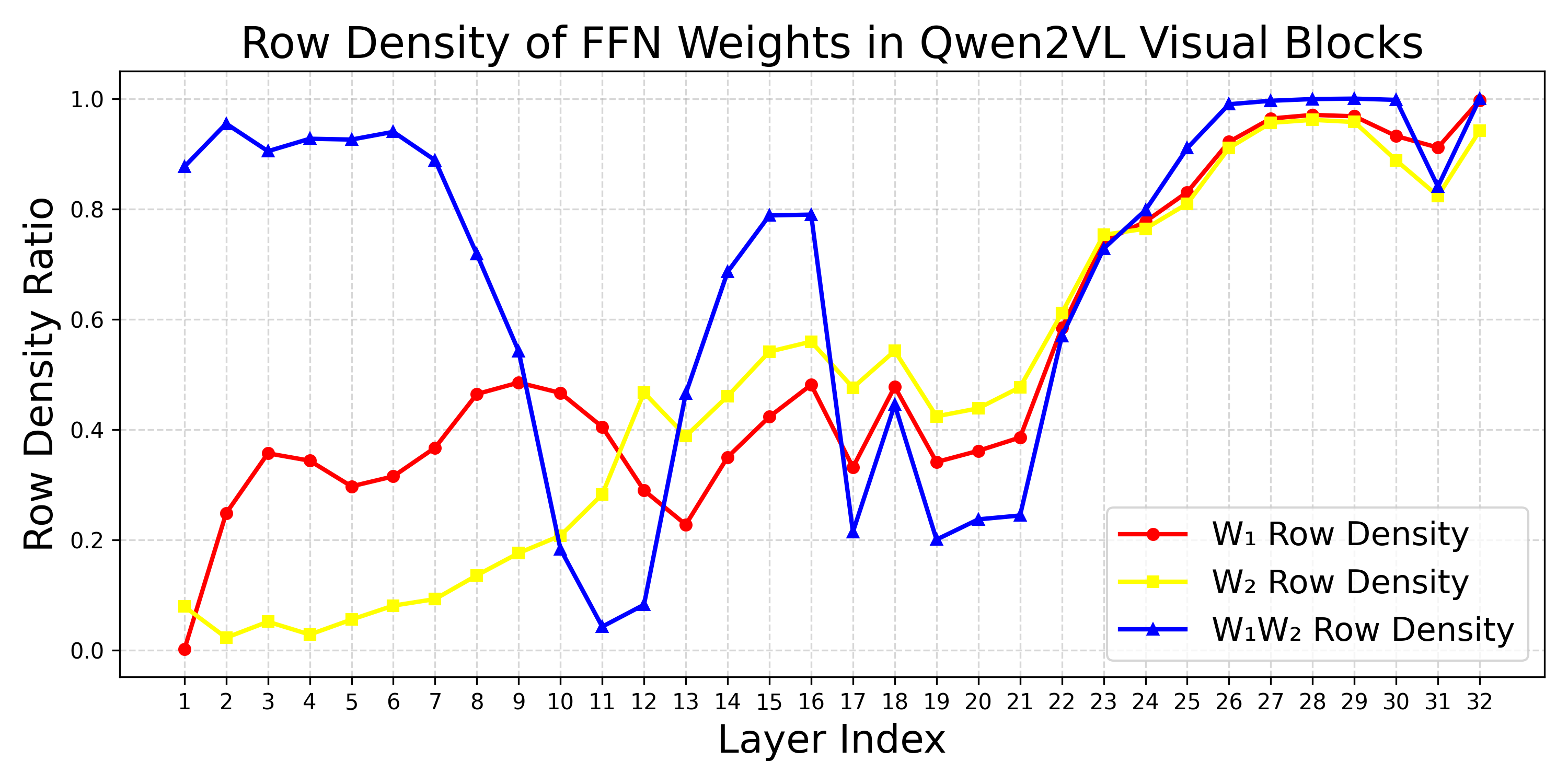}
    \caption{Row density ratio of the visual feed-forward network in Qwen2VL-7B. $W_1$ and $W_2$ denote two projection matrices, and $W_1 \times W_2$ represents their composition, as related to our method. The sparse ratio is computed via the $L_1$ norm; a row is considered sparse if its $L_1$ norm is below 5\% of the input dimension.}
    \label{ffn_sparsity}
\end{figure}

Unlike compression strategies for pure language models, the visual modules of VLMs exhibit substantial functional redundancy, stemming from both attention and feed-forward components:
(1) \textit{Redundant attention heads} occur frequently and vary in importance across layers. As shown in Figure~\ref{attention_importance}, several heads contribute negligibly to overall performance, a disparity that grows in deeper layers.
(2) \textit{Feed-forward networks (FFNs)} display notable structural sparsity, with significant variation in sparsity across layers (Figure~\ref{ffn_sparsity}). In practice, although FFNs employ nonlinear activations, their local behavior can be well-approximated by linear mappings, allowing singular value decomposition (SVD) to effectively capture dominant subspace structure.

Existing research has made progress in compressing large multimodal models through token pruning~\cite{chen2024image, jiang2025kind, meng2025plphp}, head pruning~\cite{gao2023learning, zhang2024treat, farina2024multiflow}, and FFN pruning~\cite{zhang2024treat, sengupta2025you}. Nevertheless, two core challenges remain. 
First, most current methods rely on heuristic importance measures or empirically tuned ratios, lacking a clear theoretical foundation to explicitly balance information preservation against compactness. 
Second, differential pruning ratios across modules are often determined empirically rather than derived from actual structural properties—particularly for the linear components of FFNs. 
This leads to a central question: \textit{How can we quantitatively measure information retention during compression and design head and FFN pruning strategies in a principled, information-aware manner?}

\begin{figure}[t]
    \centering
    \includegraphics[width=\linewidth, height=6cm]{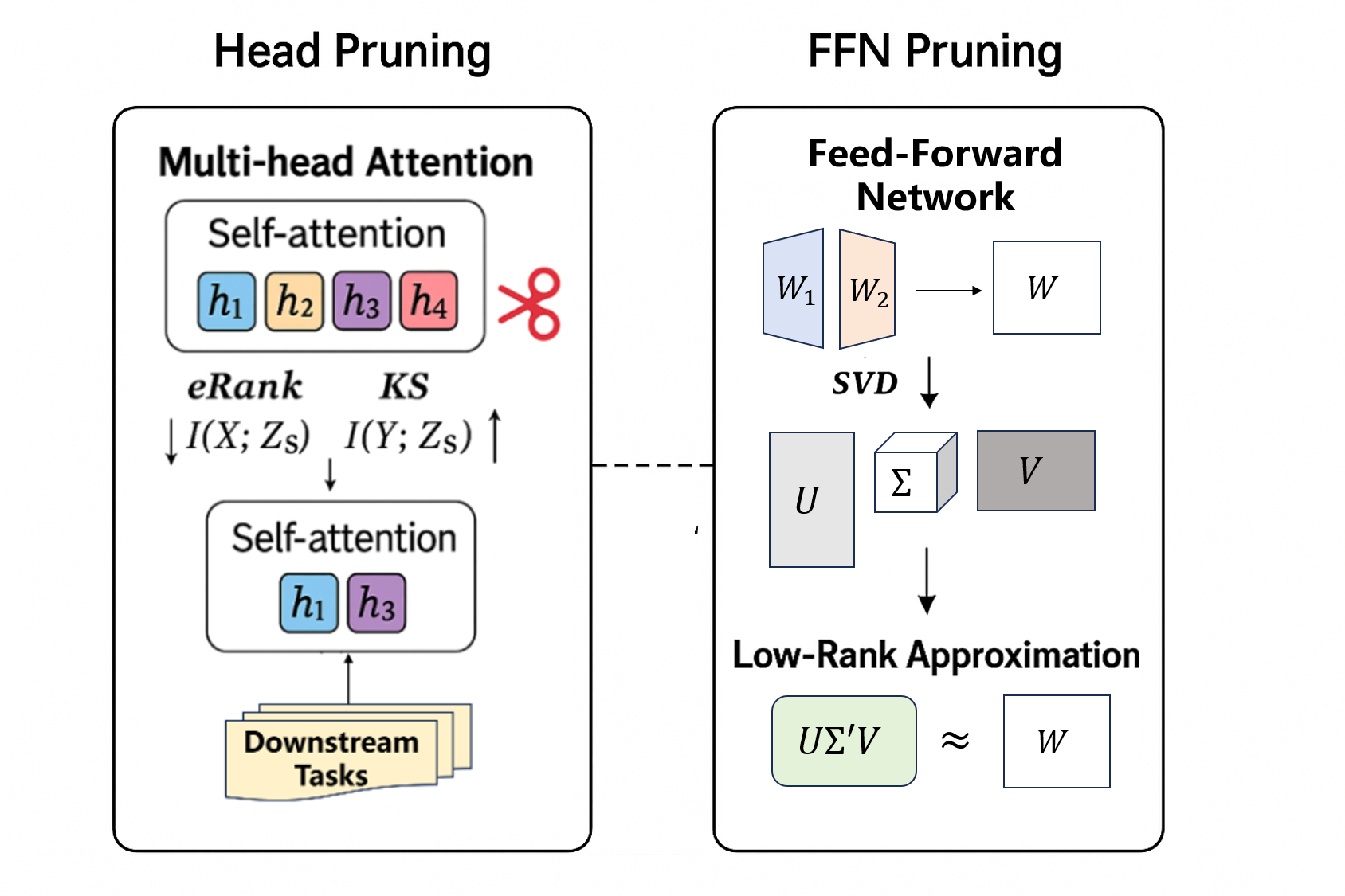}
    \caption{Overview of our method. For head pruning, eRank is used to minimize the mutual information $I(X; Z_S)$ between input $X$ and pruned representation $Z_S$, while the KS distance maximizes $I(Y; Z_S)$ between the output $Y$ and $Z_S$. For FFN pruning, we employ SVD and the Eckart--Young theorem to automatically determine the retained rank under a predefined compression target.}
    \label{method}
\end{figure}

To tackle these issues, we propose an adaptive structural pruning framework for vision-language models, grounded in unified information-theoretic principles. We employ the \textit{effective rank (eRank)} and the \textit{Kolmogorov--Smirnov (KS) distance} as complementary indicators for quantifying information retention and loss throughout the pruning process, guided by the Information Bottleneck (IB) principle~\cite{tishby2015deep}. These metrics jointly enable a principled evaluation of both attention and feed-forward modules. 

Based on this theoretical foundation, we introduce two complementary strategies:
(i) a training-based head pruning approach with learnable importance parameters optimized under an IB objective, where eRank encourages redundancy removal while KS penalizes the loss of salient information;
and (ii) a training-free FFN compression approach using SVD, where the retained rank is adaptively chosen according to a reconstruction error bound derived from the Eckart--Young theorem~\cite{eckart1936approximation}. 
Together, these techniques yield a theoretically grounded, information-aware pruning framework for VLMs.
An overview is presented in Figure~\ref{method}.

Our main contributions are as follows:
\begin{itemize}
    \item \textbf{Theoretical Contribution:} We integrate the information bottleneck principle into attention head pruning, introducing a loss that combines eRank and KS distance to provide theoretical guarantees for the trade-off between information preservation and redundancy elimination.
    \item \textbf{Algorithmic Contribution I:} We design a training-based attention head pruning algorithm grounded in the IB principle that adaptively retains informative heads while pruning redundant ones, enhancing inference efficiency.
    \item \textbf{Algorithmic Contribution II:} We propose a training-free FFN compression method based on the SVD and Eckart--Young theorem, achieving controllable low-rank approximation while maintaining model fidelity.
\end{itemize}

\section{Related Work}
\label{sec:related}

\paragraph{Vision-Language Models.}
Recent years have witnessed rapid progress in adapting Transformer-based Large Language Models (LLMs)~\cite{zhao2025surveylargelanguagemodels}, such as GPT~\cite{openai2024gpt4technicalreport} and Qwen3~\cite{yang2025qwen3}, beyond text-only understanding. 
Early dual-encoder frameworks like CLIP~\cite{radford2021learning} and ALIGN~\cite{jia2021scaling} pioneered contrastive pretraining on large-scale image–text pairs, enabling zero-shot transfer to vision tasks.
Subsequent unified architectures, including BLIP~\cite{li2022blip}, Flamingo~\cite{alayrac2022flamingo}, and BLIP-2~\cite{li2023blip}, further integrated language generation with visual reasoning. 
Recent models such as QwenVL~\cite{wang2024qwen2, bai2025qwen2} and Gemma~\cite{team2025gemma} achieve state-of-the-art results on multimodal benchmarks, marking a shift toward general-purpose vision-language understanding.

\paragraph{Model Compression in LLMs.}
To manage the rapidly growing scale of LLMs, various compression methods have been developed to enhance inference efficiency without sacrificing accuracy. 
\textit{Quantization-based} techniques~\cite{frantar2022optimal, shao2023omniquant, liu2024kivi} reduce memory footprint by lowering precision for model weights or KV caches, while \textit{pruning-based} approaches~\cite{frantar2023sparsegpt, sharma2023truth, xu2024besa, sengupta2025you} remove redundant neurons, heads, or even full layers using structured or unstructured sparsification. 
More principled alternatives leverage \textit{information-theoretic}~\cite{qin2024accurate} and \textit{low-rank approximation}~\cite{li2024svdquant, ping2024delta, wang2403svd} formulations, preserving semantics by constraining information loss during the compression.

\paragraph{Model Compression in VLMs.}
While related in spirit, compressing VLMs introduces additional complexity due to the heterogeneous nature of visual and textual modules.  
Most existing methods focus on \textit{token pruning}~\cite{shang2024llava, chen2024image, jiang2025kind, meng2025plphp}, reducing redundant visual tokens to accelerate inference, or on \textit{quantization}~\cite{li2025mbq, wang2024q}, which jointly optimizes numerical precision across modalities.
A smaller body of work explores \textit{head and FFN pruning}~\cite{gao2023learning, zhang2024treat, farina2024multiflow}, targeting structural redundancy within the visual transformer itself. 
However, these methods often depend on heuristic importance metrics or empirically tuned ratios, lacking theoretical measures of task-relevant information is preserved.

Our approach differs from existing compression frameworks by introducing a unified, information-theoretic perspective.  
We propose \texttt{InfoPrune}, which combines the \textit{effective rank (eRank)} and the \textit{Kolmogorov--Smirnov (KS) distance} under the Information Bottleneck principle to jointly guide attention and FFN pruning.  
This formulation provides a principled criterion for information-aware structural compression, bridging the gap between empirical heuristics and theoretically grounded model reduction.
\section{Background}
\subsection{Transformer Architecture}
We denote a two-layer transformer model, where the first layer is a self-attention mechanism with a softmax activation, and two fully connected layers with weight matrices $W_1 \in \mathbb{R}^{d \times m}$ and $W_2 \in \mathbb{R}^{m \times d}$. The block output can be written as:
\begin{equation}
    \begin{aligned}
    \nonumber f&=\phi(ZW_1)W_2\\
    \nonumber Z& =\text{MultiHead}(Q, K, V), \\
    &= \text{Concat}(\text{head}_1, \ldots, \text{head}_h), 
    \end{aligned}
\end{equation}
where $\text{head}_i = \text{Attention}(Q W_i^Q, K W_i^K, V W_i^V)$ ,for $i = 1, 2, \ldots, h$ and $\phi$ is a non-linear function. The parameters of the attention layer are defined as
$W_i^Q \in \mathbb{R}^{d\times d_k}$
$W_i^K \in \mathbb{R}^{d \times d_k}$
$W_i^V \in \mathbb{R}^{d\times d_v}$.

\subsection{Information Bottleneck}
The idea behind the Information Bottleneck (IB) is to extract a feature, denoted as $Z$, that extracts all the effective information required for the task $Y$ and discards redundant information $X$. Mathematically, this is formulated as:
\begin{align}
\min  I(X;Z) - \beta I(Y;Z),
\end{align}
where $\beta$ is a Lagrange multiplier that controls the balance between information compression and task relevance. 
The IB principle provides the theoretical foundation for our pruning objective, where we explicitly model and control the retention of informative structures during compression.
\subsection{SVD and Low-Rank Approximation}
Singular Value Decomposition (SVD) provides an essential tool for analyzing linear layers and constructing compact parameterizations. 
Given a matrix $W \in \mathbb{R}^{m \times n}$, the SVD is defined as $W = U \Sigma V^\top$, where 
$U \in \mathbb{R}^{m \times m}$ and $V \in \mathbb{R}^{n \times n}$ are orthonormal matrices, and 
$\Sigma = \mathrm{diag}(\sigma_1, \sigma_2, \ldots, \sigma_r)$ contains singular values in descending order 
($\sigma_1 \ge \sigma_2 \ge \cdots \ge \sigma_r > 0$), where $r=\mathrm{rank}(W)$.
According to the Eckart--Young--Mirsky theorem, the truncated representation 
\[
\widehat{W} = \sum_{i=1}^{r'} \sigma_i u_i v_i^\top, \quad r' \le \min\{m,n\},
\]
yields the optimal rank-$r'$ approximation of $W$ under the Frobenius norm. 
This result provides a theoretical basis for low-rank compression of feed-forward layers in VLMs.

\subsection{Effective Rank}
The \textit{effective rank (eRank)} quantifies the intrinsic dimensionality of a matrix based on the entropy of its singular value distribution. 
For $A \in \mathbb{R}^{d \times N}$ with singular values $\{\sigma_i\}_{i=1}^Q$, where $Q=\min\{d,N\}$, the eRank is defined as:
\begin{equation}
\mathrm{eRank}(A) = 
\exp\left(
    -\sum_{i=1}^{Q} 
    p_i \log p_i
\right), 
\quad p_i = \frac{\sigma_i}{\sum_{j=1}^{Q} \sigma_j}.
\end{equation}

Unlike the conventional matrix rank, which depends on arbitrary thresholds, eRank captures the effective number of dominant singular directions that contribute to the information content of $A$. 
This notion is crucial for our method, as it quantitatively reflects redundancy and helps guide adaptive structural pruning.

\section{Method}

\subsection{Head Pruning with Information Bottleneck}

\begin{figure}
    \centering
    \includegraphics[width=1\linewidth, height=6cm]{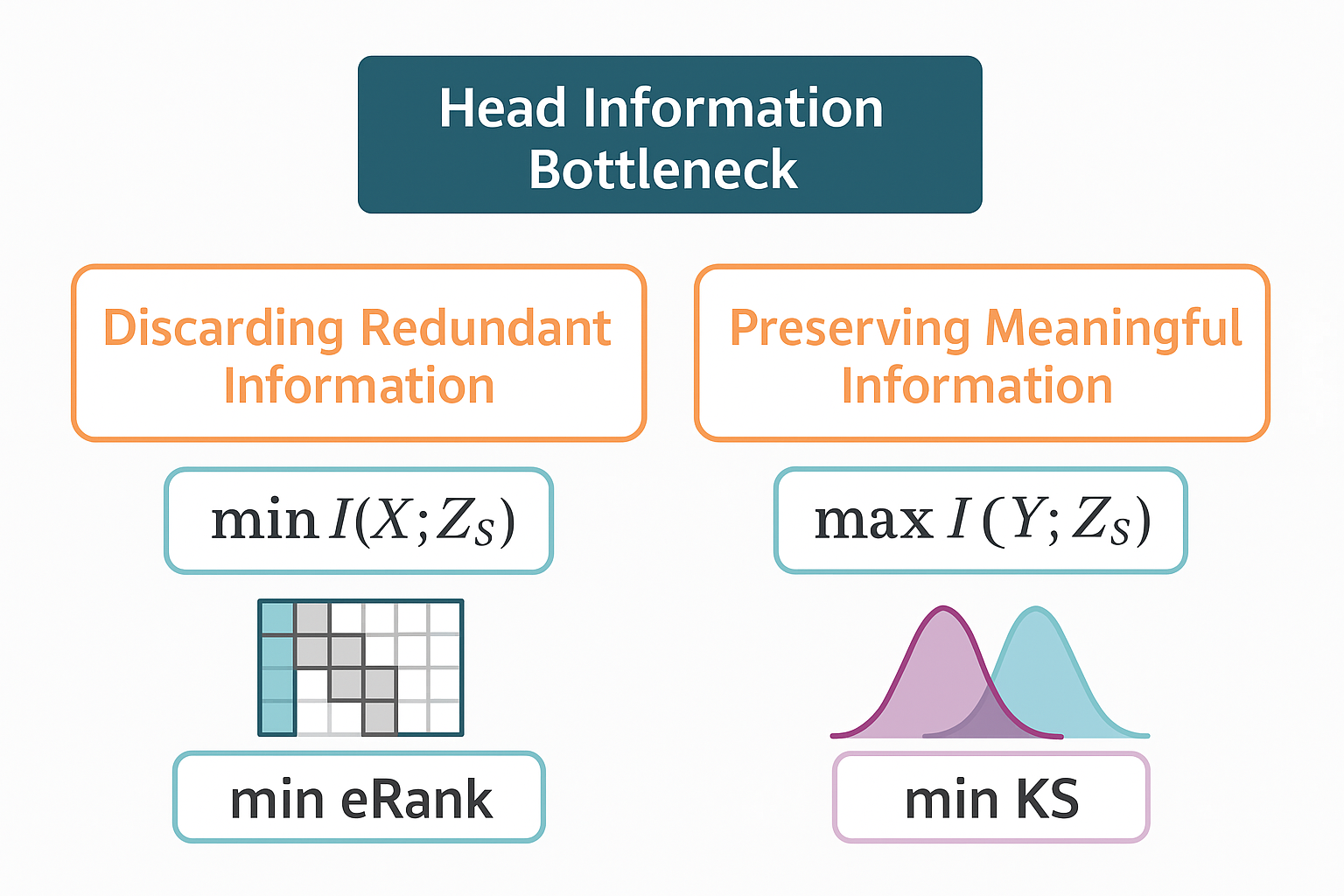}
    \caption{A visualized description for head pruning. The method is divided into two main strategies: discarding redundant information and preserving meaningful information. On the left, the goal is to minimize \( I(X; Z_S) \) to reduce redundancy, which is further refined by minimizing the eRank of the pruned model. On the right, the focus is on maximizing  \( I(Y; Z_S) \) to retain meaningful information, which involves SVD to select the first \( k \) largest sigma and minimize the KS distance.}
    \label{head_pruning}
\end{figure}

We found many similarities between the Information Bottleneck (IB) \cite{tishby2000information} theory and attention mechanisms. In the attention mechanism, attention reweights the input data and gives higher weight to the most effective information. The core idea of information bottleneck theory is to strike an optimal balance between compressing input information and preserving information relevant to the output. It views the model's learning process as maximizing task-relevant representation while minimizing irrelevant or redundant input features. The essence of both approaches is to retain only the most critical information. Therefore, we apply the information bottleneck theory to the attention mechanism of multiple heads to prune the opposite head, in Figure~\ref{head_pruning}.

Let the indices of the retained attention heads after pruning be \(S \subseteq \{1, 2, \ldots, h\}\), where \(|S| < h\), and the theory suggests minimizing:
\begin{align}
\min_{Z_S \subseteq Z} \quad & I(X; Z_S) - \beta I(Y; Z_S).
\end{align}

The goal is to select the optimal combination of attention heads from a total of $h$ heads to form $Z_S$, such that the following conditions are satisfied after pruning:
\begin{itemize}
    \item The intermediate output $Z_S$ discards more redundant information: $I(X; Z_S) < I(X; Z)$.
    \item The meaningful information in the intermediate output $Z_S$ remains unchanged: $I(Y; Z_S) \approx I(Y; Z)$.
\end{itemize}

\subsubsection{Discarding Redundant Information $I(X; Z_S)$}

To minimize redundancy in the pruned output matrix, we aim to ensure that the difference in mutual information \( I(X; Z_S) - I(X; Z) \) is negative, indicating that the pruned model retains significant information while discarding redundancy, which leads to the use of effective rank. A detailed proof is provided in Appendix A.

\begin{theorem}[\textbf{Head Pruning and eRank Optimization}]
\label{theo-e}
Our objective is to minimize the redundancy in the pruned model. It can be proved that the difference in mutual information \( I(X; Z_S) - I(X; Z) \) is equivalent to the difference in eRank between the pruned and original intermediate outputs, i.e.
\begin{equation}
\label{eq_1}
    \begin{aligned}
        & \min I(X; Z_S) - I(X; Z) \\
        \Longleftrightarrow & \min \mathrm{eRank}(Z_S) - \mathrm{eRank}(Z).
    \end{aligned}
\end{equation}
\end{theorem}

Therefore, minimizing the effective rank difference between the pruned and original matrices ensures that the pruned representation \( Z_S \) retains the most informative components while discarding redundancy. The corresponding loss term is defined as:
\begin{equation}
\label{Lerank}
    \mathcal{L}_{\mathrm{eRank}} = \mathrm{eRank}(Z_S) - \mathrm{eRank}(Z).
\end{equation}

\subsubsection{Preserving Meaningful Information $I(Y; Z_S)$}

To ensure that the pruned output matrix retains more meaningful information, we aim to minimize the difference in mutual information $I(Y; Z_S) – I(Y; Z)$. Using singular value decomposition (SVD), we extract the singular values of the original matrix $Z$, retaining the larger, significant values and discarding the smaller, redundant ones. This low-rank approximation preserves the most relevant information for predicting task $Y$. Guided by information bottleneck, we quantify the discrepancy between the singular value distributions of the original and pruned matrices using Kolmogorov-Smirnov (KS) distance, which is related to the SVD theorem. A detailed proof is provided in the supplementary material.

\begin{theorem}[\textbf{Head Pruning and KS Distance}]
\label{theo-k}
Let the original matrix \( Z \) have the singular value decomposition \( \Sigma_Z = \text{diag}(\sigma_1, \sigma_2, \ldots, \sigma_{r_1}) \). By retaining only the \( k \) largest singular values, we derive the pruned singular value matrix \( \Sigma_{Z_S} = \text{diag}(\sigma_1, \sigma_2, \ldots, \sigma_k) \). To assess the distributional discrepancy between the spectra, we compute the KS distance, defined as:
\begin{align*}
\mathcal{D} = \sup_{x} \left| F_{1,r_1}(x) - F_{2,k}(x) \right|,
\label{ks}
\end{align*}
where \( F_{1,r_1} \) and \( F_{2,k} \) denote the empirical distribution functions of \( \{\sigma_i\}_{i=1}^{r_1} \) and \( \{\sigma_i\}_{i=1}^{k} \) respectively. To maximize the retention of significant information, it is essential to minimize the pruning rate, effectively aiming to minimize the KS distance, which can be mathematically described as:
\begin{align*}
\min & \  I(Y; Z_S) - I(Y; Z) \Longleftrightarrow \min KS.
\end{align*}
\end{theorem}

Accordingly, the KS-based loss term is defined as:
\begin{equation}
\label{LKS}
    \mathcal{L}_{\mathrm{KS}} = \sup_{x} \left| F_{1,r_1}(x) - F_{2,k}(x) \right|,
\end{equation}
which encourages the pruned representation \( Z_S \) to preserve the spectral characteristics of the original matrix \( Z \), thereby retaining meaningful information essential for the prediction of \( Y \).

\subsubsection{A Train-based Head Pruning Method}

From the perspective of information bottleneck theory, we propose a dynamic head pruning method that adjusts each attention head's importance during training. This approach reduces redundancy without affecting the model's ability to capture key cross-modal interactions. We introduce a learnable importance parameter $\zeta_{l,h}$ for the $h$-th attention head in the $l$-th layer, initialized to a large value, whose Sigmoid value is close to 1, ensuring no initial pruning.

Motivated by Theorem~\ref{theo-e} and Theorem~\ref{theo-k}, we design a comprehensive loss function that considers task loss, pruning sparsity, and redundancy minimization. The loss function is expressed as follows:
\begin{equation}
\mathcal{L} = \mathcal{L}_{\mathrm{task}} + \alpha \|\zeta\|_1 + \beta \mathcal{L}_{\mathrm{eRank}} + \gamma \mathcal{L}_{\mathrm{KS}},
\end{equation}
where $\mathcal{L}_{\mathrm{task}}$ represents the task-related loss term; $\|\zeta\|_1$ is the $\ell_1$ regularization term for all $\zeta_{l,h}$, aimed at encouraging sparsely selecting; $\mathcal{L}_{\mathrm{eRank}}$ is defined in Equation (\ref{Lerank}); and $\mathcal{L}_{\mathrm{KS}}$ is defined in Equation (\ref{LKS}). The hyperparameters $\alpha$, $\beta$, and $\gamma$ control the weight of each loss term in the overall loss.

During training, we dynamically adjust the weight $\zeta_{l,h}$ to achieve the optimal pruning strategy. After training, based on the final importance parameter vector $\zeta$, we apply Sigmoid to map the values to the range [0, 1] and prune heads with the mapped value according to a predefined pruning threshold $z \in (0, 1)$. A detailed pruning procedure is presented in Algorithm~\ref{head-prune}.

\paragraph{Complexity Analysis.} The training phase optimizes the learnable weights $\zeta_{l,h}$ of each attention head. Assuming a total number of training steps of $T$, each step involves $O(LH)$ updates, resulting in a complexity of $O(TLH)$. The pruning phase traverses all $LH$ attention heads for selection, with a complexity of $O(LH)$. Therefore, the overall time complexity is $O(TLH)$, while the space complexity is dominated by the vector $\zeta$, which is $O(LH)$, making it relatively lightweight in practice.

This pruning method adaptively selects the heads to retain in each layer, while ensuring effective retention of cross-modal information and minimal redundancy under the guarantee of information bottleneck theory.

\begin{algorithm}[htbp]
	\small
	\caption{Train-based Attention Head Pruning}
	\label{head-prune}
	\textbf{Input:} Transformer model with $L$ layers and $H$ attention heads each layer; attention head sets $S_H$; pruning threshold $z \in (0,1)$; loss weights $\alpha, \beta, \gamma$. \\
	\textbf{Output:} Selected attention heads $S$.
    
	\begin{algorithmic}[1]
		\STATE $\zeta_{l,h} \gets \text{large value}$, $S \gets \emptyset$
		\FOR{each training step}
			\STATE $\mathcal{L} \gets \mathcal{L}_{\mathrm{task}} + \alpha \|\zeta\|_1 + \beta \mathcal{L}_{\mathrm{eRank}} + \gamma \mathcal{L}_{\mathrm{KS}}$
			\STATE Backpropagate and update $\zeta_{l,h}$
		\ENDFOR
		\FOR{each head $ \in S_H$}
			\IF{$\text{Sigmoid}(\zeta_{l,h}) \geq z$}
				\STATE $S \gets S \cup \{(\ell, h)\}$
			\ENDIF
		\ENDFOR
		\STATE \textbf{return} $S$
	\end{algorithmic}
\end{algorithm}

\subsection{FFN Compression with SVD}

While attention head pruning is designed to be trainable and
adaptive to task-specific importance, the pruning of Feed-Forward
Networks (FFNs) follows a different principle. In vision-language
models, each FFN typically adopts a non-linear
activation between two linear layers:
\[
  y = W_2 \phi(W_1 x),
\]
where $\phi(\cdot)$ denotes a non-linear function, e.g. the SwiGLU function in Qwen2-VL. This nonlinearity
introduces expressive flexibility but also complicates direct analysis.

To ensure a tractable and training-free compression scheme, we
approximate the nonlinear mapping by its locally linear form around
the typical activation region of $\phi$. Specifically, by performing
a first-order Taylor expansion of $\phi(W_1 x)$ around input $x_0$,
we obtain a linearized representation:
\[
  y \approx W_2 \, \text{diag}(\phi'(W_1 x_0)) \, W_1 (x - x_0) + b,
\]
where $\text{diag}(\phi'(W_1 x_0))$ acts as a data-dependent scaling term.
This perspective allows us to treat each FFN as an effective linear map
in a local subspace, justifying the use of Singular Value Decomposition
(SVD) for analyzing and compressing its dominant directions.

Under this assumption, we compute the SVD of the composite weight:
\[
  W \approx W_2 \, \text{diag}(\phi'(W_1 x_0)) \, W_1 = U \Sigma V^\top,
\]
where $U, V \in \mathbb{R}^{d \times d}$ are orthogonal matrices, and
$\Sigma = \mathrm{diag}(\sigma_1, \dots, \sigma_r)$ contains singular values
ordered in descending magnitude $\sigma_1 \ge \dots \ge \sigma_r > 0$.
We then retain the top-$k$ singular components that satisfy a
reconstruction error constraint:
\begin{align}
  \frac{\|W - \widehat{W}_k\|_F}{\|W\|_F} \le \epsilon,
  \label{error}
\end{align}
where $k$ is the target rank after pruning, and $\| \cdot \|_F$ denotes the Frobenius norm. To ensure minimal information loss, we aim to keep $\epsilon$ as small as possible.

The following theorem provides a principled way to determine the minimal rank $k$ needed to meet a given error threshold.

\begin{theorem}[\textbf{Adaptive FFN Pruning}]
\label{theo-ey}
Given a matrix $W$ with singular values $\sigma_1 \ge \sigma_2 \ge \cdots \ge \sigma_r$, the smallest rank $k$ such that the reconstruction error satisfies Equation (\ref{error}) 
is the minimal integer $k$ satisfying:
\begin{equation}
\label{FFN-pruning}
    \sum_{i = 1}^k \sigma_i^2 \geq (1 - \epsilon^2) \sum_{i=1}^r \sigma_i^2.
\end{equation}
\end{theorem}

A proof of Theorem \ref{theo-ey} is in Appendix C. From Equation (\ref{FFN-pruning}), we can conclude that once the singular values are obtained, it is possible to directly determine the retained rank $k$ that satisfies the error threshold condition for a given $\epsilon$. Specifically, we select the smallest $k$ such that the sum of the squared top-$k$ singular values accounts for at least a $(1 - \epsilon^2)$ proportion of the total squared singular values. 

The detailed pruning procedure based on this criterion is presented in Algorithm~\ref{alg:adaptive-prune}.

\begin{algorithm}
	\small % Reduce font size to fit slide width if needed
	\caption{Layer-wise FFN Pruning via Singular Value Decomposition}
	\label{alg:adaptive-prune}
	\textbf{Input:} Weight matrices $W_1 \in \mathbb{R}^{d \times m}$, $W_2 \in \mathbb{R}^{m \times d}$; similarity threshold $\epsilon \in (0, 1]$. \\
	\textbf{Output:} Compressed matrices $\hat{W_1} \in \mathbb{R}^{d \times k}$, $\hat{W_2} \in \mathbb{R}^{k \times d}$, where $k$ is the adaptive rank.
	\begin{algorithmic}[1]
		\STATE $W \gets W_1 W_2$  
		\STATE $[U, \Sigma, V^\top] \gets \text{SVD}(W)$, where $\Sigma = (\sigma_1, \ldots, \sigma_r)$
		\STATE $S \gets \sum_{i=1}^r \sigma_i^2$; $T \gets (1 - \epsilon^2) \cdot S$; $E \gets 0$
		\FOR{$k = 1$ to $r$}
			\STATE $E \gets E + \sigma_k^2$
			\IF{$E \ge T$} 
                \STATE \textbf{break}
		      \ENDIF
        \ENDFOR
		\STATE $\hat{U} \gets U[:, 1\!:\!k];\ \hat{\Sigma} \gets \Sigma[1\!:\!k,\ 1\!:\!k];\ \hat{V} \gets V[:, 1\!:\!k]$
		\STATE $\hat{W}_1 \gets \hat{U} \hat{\Sigma}^{1/2};\ \hat{W}_2 \gets \hat{\Sigma}^{1/2} \hat{V}^\top$
		\STATE \textbf{return} $\hat{W}_1, \hat{W}_2$
	\end{algorithmic}
\end{algorithm}

\paragraph{Complexity Analysis.} The time complexity of Algorithm~\ref{alg:adaptive-prune} is $O(d^2 m + d^3)$, where the first term comes from computing $W = W_1 W_2$, and the second from performing full SVD on $W \in \mathbb{R}^{d \times d}$. The search for the minimal rank $k$ adds only $O(d)$ overhead. The space complexity is $O(d^2)$, dominated by storing $W$ and its SVD components. 
\section{Experiments}

\begin{table*}[!t]
\centering
\small  % font less than 9pt
\renewcommand{\arraystretch}{1.25} 
\renewcommand{\tabcolsep}{2.5pt}
\begin{tabular}{lccccccc}
    \toprule
    \textbf{Method} & \textbf{Vision FLOPs (T)} & \textbf{Forward Latency (ms)} & \textbf{Speed-up Ratio} & \textbf{VQAv2} & \textbf{TextVQA} & \textbf{POPE} & \textbf{MME}  \\
    \midrule
    Baseline & 24.33 & 650.37 & 1.00 & 81.82 & 74.37 & 78.95 & 82.78  \\
    \midrule
    \multicolumn{8}{c}{\textbf{Attention Head Pruning Methods}} \\ 
    % \midrule
    Random & 52.84 & 346.23 & 1.88 & 57.58 & 66.67 & 53.85  & 51.61  \\
    YOPO & \cellcolor{green!20}\textbf{11.68} & 730.75 & 0.89 & \cellcolor{green!20}\textbf{63.56} & \cellcolor{green!20}\textbf{73.74} & 74.11 & \cellcolor{green!20}\textbf{80.33} \\
    \textbf{InfoPrune (Ours)} & 48.02 & \cellcolor{green!20}\textbf{328.94} & \cellcolor{green!20}\textbf{1.98} & 62.63 & 73.60 & \cellcolor{green!20}\textbf{74.79} & 76.67 \\
    \midrule
    \multicolumn{8}{c}{\textbf{FFN Pruning Methods}} \\
    % \midrule
    Random & 38.67 & 344.35 & 1.89 & \cellcolor{green!20}\textbf{62.63} & 31.33 & 55.41 & 54.30  \\
    PruneNet & \cellcolor{green!20}\textbf{17.19} & 429.54 & 1.51 & 40.53 & 34.67 & 47.68 & 50.85 \\
    \textbf{InfoPrune (Ours)} & 37.90 & \cellcolor{green!20}\textbf{330.65} & \cellcolor{green!20}\textbf{1.97} & 59.38 & \cellcolor{green!20}\textbf{40.94} & \cellcolor{green!20}\textbf{60.87} & \cellcolor{green!20}\textbf{58.56} \\
    \bottomrule
\end{tabular}
\caption{Comparison of different pruning methods across benchmarks. The methods are categorized into two groups: attention head pruning and FFN pruning. In our method (InfoPrune), we compress attention heads using a threshold of $z = 0.5$ (69\% head pruning rate) and approximate FFNs with $\epsilon = 0.001$ (76\% pruning rate). The Vision FLOPs and forward latency values represent averages across all visual layers, and the Speed-up Ratio is calculated based on the forward latency. Additionally, since MME and POPE do not require training, we utilize the model post-trained on VQAv2 for evaluation.}
\label{tab:pruning}
\end{table*}

\subsection{Experiment Setups}
This work primarily focuses on pruning computations within the vision module of the Qwen2 series vision-language models, particularly the visual model Qwen2VL-7B \cite{wang2024qwen2}. The vision module of Qwen2VL-7B consists of 32 stacked Qwen2VLVisionBlocks. 16 attention heads are contained in each block, the FFN in which has an input dimension of 1280 and a hidden dimension of 5120. 

In head pruning, we consider the hyperparameters $\alpha = 0.01, \beta = 0.01, \gamma = 0.1$, use the AdamW \cite{loshchilov2017decoupled} optimizer with a learning rate of $1e^{-2}$ and 7 epoches. During training, our learning rate will gradually decrease according to the predefined learning rate decay strategy. All the experiments were performed on a single NVIDIA A800-120GB GPU. Each data point represents the average value obtained from multiple trials. During tuning, InfoPrune requires about 60 GB of GPU memory, The effective tuning time is approximately 5 hours for about 1000 epochs on a single GPU.

% description for hyper-parameters
\paragraph{Benchmarks.} To evaluate the performance of pruned models, we test their zero-shot capabilities on various vision-language benchmarks. We use FLOPs and forward latency to measure computational efficiency, while focusing on tasks such as visual reasoning, visual description, and visual extraction. Specifically, \textit{VQAv2} \cite{goyal2017making} assesses the model's ability to perform open-ended image question answering by combining visual perception and common sense. \textit{TextVQA} \cite{singh2019towards} evaluates the model's capability to understand and answer questions based on text within images. \textit{POPE} \cite{Li-hallucination-2023} targets object hallucination detection in visual models. \textit{MME} \cite{fu2024mmecomprehensiveevaluationbenchmark} examines logical reasoning under visual information interaction through yes/no questions. These benchmarks provide a comprehensive evaluation of our proposed method's performance in visual domain tasks.

\paragraph{Baselines.}  To evaluate the effectiveness of our proposed pruning framework, we compare it with several representative baseline methods in the domain of vision-language model compression.

These include head pruning baselines like YOPO \cite{zhang2024treat}, and FFN pruning methods like PruneNet \cite{sengupta2025you}. We also include a random pruning strategy and the original unpruned model as references. These baselines cover training-free and empirical pruning approaches, serving as strong representatives of current mainstream practices in VLM compression.

\paragraph{Ablation Studies.} In order to validate the effectiveness of our method, we designed three types of ablation experiments. First, for the head pruning method, we performed an ablation study on the loss function components, removing eRank and KS distance to verify the role of information retention and redundancy removal in the pruning process, and further explore the contribution of different loss function components to the model's pruning effectiveness. Additionally, we conducted an ablation experiment on pruning ratios for both head pruning and FFN pruning, comparing the performance changes under different pruning ratios to study the balance between pruning extent, computational efficiency, and model performance. These ablation experiments help to deepen our understanding of the necessity of each design choice in our method and validate its effectiveness in practical applications.

\subsection{Comparison Results}
\begin{figure}[t]
    \centering

    \begin{subfigure}{1.0\linewidth}
        \centering
        \includegraphics[width=\linewidth]{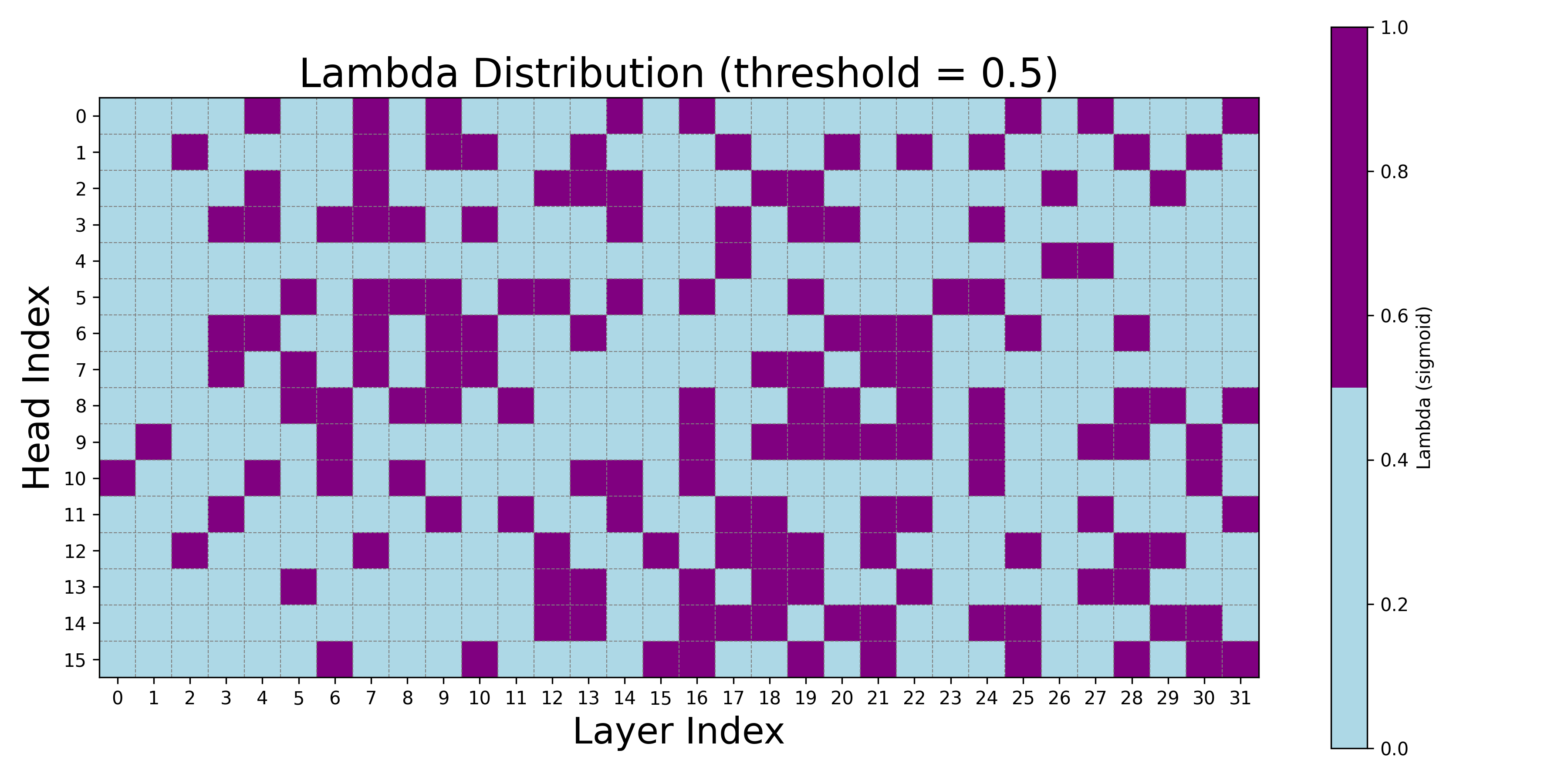}
        \caption{Training on the TextVQA, the results of head pruning with threshold 0.5. Lightblue indicates pruned while indicates retained.}
        \label{fig:lambda}
    \end{subfigure}

    \begin{subfigure}{1.0\linewidth}
        \centering
        \includegraphics[width=\linewidth]{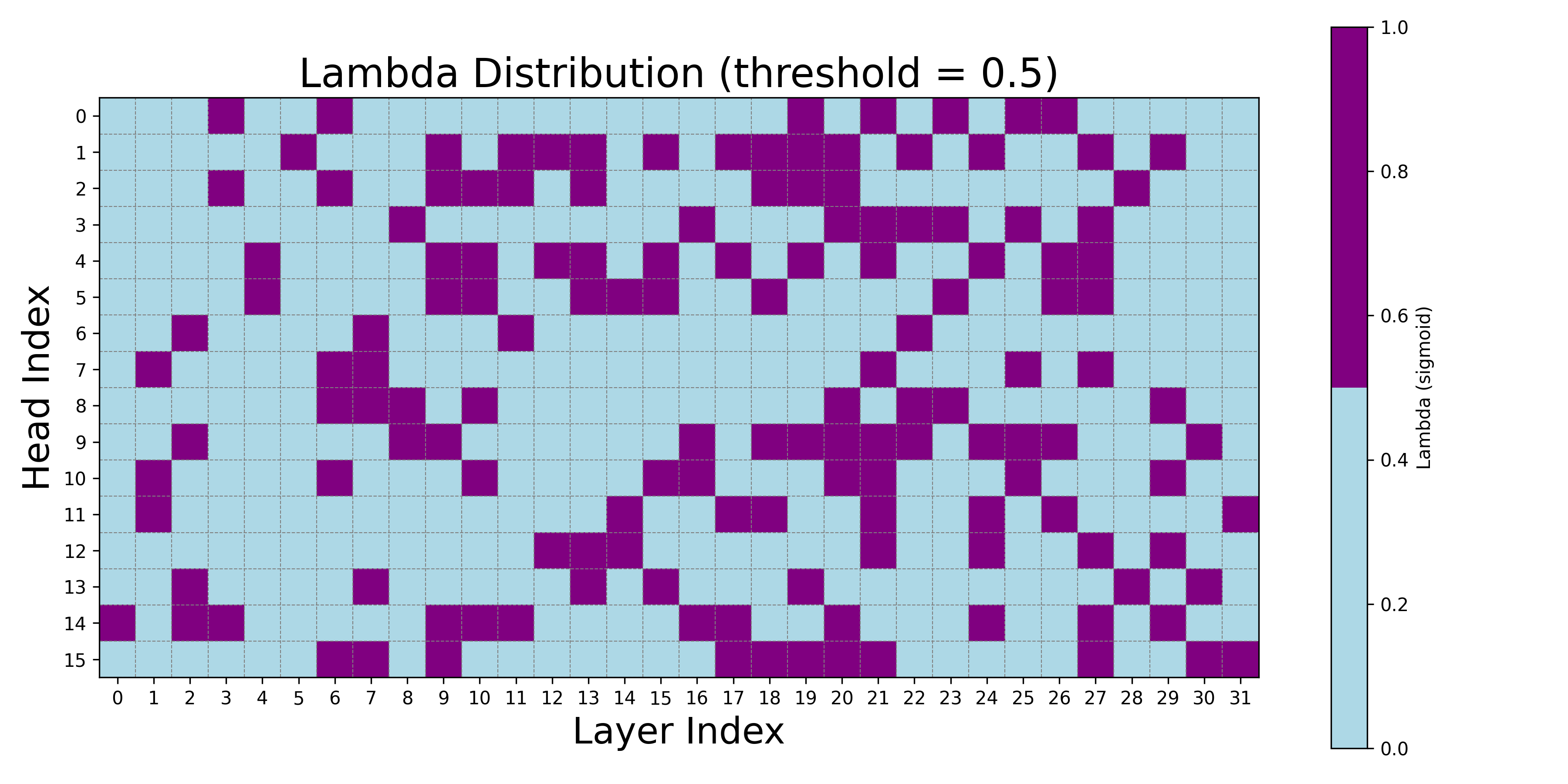}
        \caption{Same visualization in the VQAv2 setting.}
        \label{fig:lambda-v}
    \end{subfigure}

    \caption{Visualization of head pruning results across different datasets.}
    \label{fig:lambda-both}
\end{figure}

\begin{figure}[!t]
    \centering
    \includegraphics[width=1.0\linewidth]{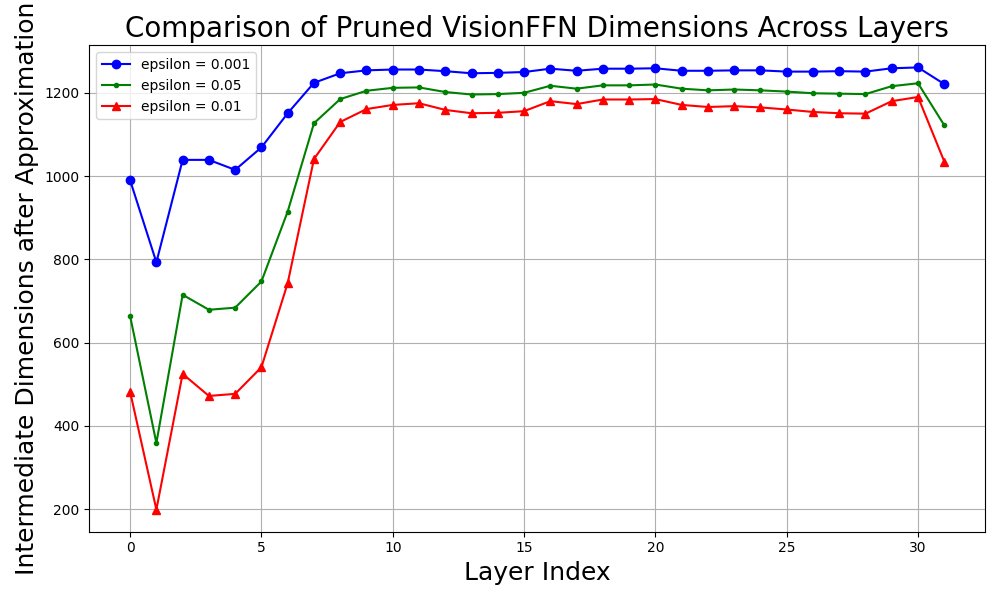}
    \caption{The intermediate dimensions of the vision FFNs in Qwen2VL-7B after low-rank approximation, with $\epsilon$ values of 0.001, 0.05, and 0.01, exhibit average dimensions of 1200, 1010, and 1087, respectively.}
    \label{pruned_mlp}
\end{figure}

Figure~\ref{fig:lambda-both} and Figure~\ref{pruned_mlp} show the pruning results, revealing that attention heads are pruned more lightly in intermediate layers, while MLP dimensions are reduced more in shallow layers.

Table~\ref{tab:pruning} reports the zero-shot performance of Qwen2VL-7B under different pruning strategies. From a computational efficiency perspective, InfoPrune achieves the lowest inference time despite a relatively high FLOPs usage. In contrast, although YOPO achieves the lowest FLOPs (11.68T) in Attention Head pruning, its inference time increases to 730.75ms, demonstrating that simply reducing FLOPs does not necessarily lead to speedup.

\subsection{Ablation Results} 

\paragraph{Loss Function Components.}

\begin{table}[h]
\centering
\small
\renewcommand{\arraystretch}{1.25} 
\begin{tabular}{l c c c}
\hline
$z$ & \textbf{Method} & \textbf{Pruning Ratio} & \textbf{TextVQA} \\
\hline
Baseline & -- & -- & 74.37   \\
\hline
\multirow{2}{*}{0.2} 
    & eRank + KS & 2.93 & 73.85 \\
    & only KS    & 5.66 & 73.11 \\
\hline
\multirow{2}{*}{0.3} 
    & eRank + KS & 4.30 & 74.41   \\
    & only KS    & 16.60 & 66.67 \\
\hline
\multirow{2}{*}{0.4} 
    & eRank + KS & 16.99 & 72.73 \\
    & only KS    & 38.28 & 67.68 \\
\hline
\multirow{2}{*}{0.5} 
    & eRank + KS & 69.14 & 73.60 \\
    & only KS    & 68.16 & 58.93 \\
\hline
\end{tabular}
\caption{Results when using eRank + KS and only KS as components of the loss function under different pruning thresholds. When KS was removed and only eRank was retained, the loss function required a very large initial learning rate to generate effective gradients, so this strategy was abandoned.}
\label{ks_vs_erank}
\end{table}

To analyze the impact of different loss function components, we compared eRank + KS with only KS at various pruning ratios. We found that using only eRank required a very high initial learning rate for effective gradient updates, making it less viable. As shown in Table~\ref{ks_vs_erank}, at smaller pruning ratios, eRank + KS slightly underperforms the baseline in TextVQA accuracy but significantly outperforms KS alone. This suggests that the combination of eRank and KS effectively preserves information and reduces redundancy.

\paragraph{Pruning Ratios.}

Table~\ref{diff-z} shows the results of attention head pruning with varying pruning threshold $z$. As $z$ increases, the visual FLOPs gradually decrease, the pruning ratio increases, and the model performance also declines, but the decrease is relatively small. This suggests that, in training-based head pruning, gradually removing redundant attention heads can effectively reduce computational complexity without causing significant performance loss.

Table~\ref{diff-ep} presents the results of FFN pruning with varying $\epsilon$ values. Similar to attention head pruning, FFN pruning also leads to a reduction in visual FLOPs. However, the key difference is that the pruning ratio increases less dramatically, but the impact on model performance is more complex. As $\epsilon$ decreases and the pruning ratio increases, there is a more noticeable decline in performance. This indicates that FFN pruning is highly sensitive to changes in the pruning ratio, and excessive low-rank compression may result in the loss of important information, negatively impacting model performance.

\begin{table}[t]
\centering
\small
\renewcommand{\arraystretch}{1.25} 
\begin{tabular}{lccc}
    \toprule
    $z$  & \textbf{Vision FLOPs (T)} & \textbf{Pruning Ratio (\%)} & \textbf{VQAv2}   \\
    \midrule
    0.2 & 52.85 & 2.93 &  69.70 \\
    0.3 & 52.67 & 4.30 &  67.68 \\
    0.4 & 52.04 & 16.99 & 67.65 \\
    0.5 & 51.52 & 69.14 & 62.63 \\
\bottomrule
\end{tabular}
\caption{The effect of different pruning ratios on vision FLOPs and model performance (VQAv2) with varying values of $z$ in head pruning.}
\label{diff-z}
\end{table}

\begin{table}[t]
\centering
\small
\renewcommand{\arraystretch}{1.25} 
\begin{tabular}{lccc}
    \toprule
    $\epsilon$  & \textbf{Vision FLOPs (T)} & \textbf{Pruning Ratio (\%)} & \textbf{VQAv2}   \\
    \midrule
    0.01 & 34.02 & 50.75 & 54.30 \\
    0.005 & 35.32 & 49.25 & 56.29 \\
    0.001 & 37.90 & 47.76 & 59.38 \\
    \bottomrule
\end{tabular}
\caption{The effect of different pruning ratios on vision FLOPs and model performance (VQAv2) at varying $\epsilon$ values in FFN pruning.}
\label{diff-ep}
\end{table}

\subsection{Discussions} 
InfoPrune consistently outperforms baseline methods in balancing efficiency and performance. In attention head pruning, it achieves a high pruning ratio (up to 69\%) with minimal accuracy drop, demonstrating effective redundancy removal while preserving task-relevant information. In contrast, random and existing learning-based methods either sacrifice too much accuracy or fail to offer meaningful speed-up.

Ablation results show that combining eRank and KS distance in the loss function is crucial. KS alone leads to aggressive and harmful pruning, while eRank guides the model to retain informative components. Additionally, head pruning proves more tolerant to high compression than FFN pruning, which is more sensitive to low-rank approximation. These findings underscore the value of InfoPrune’s information-theoretic and spectral design in compressing large vision-language models effectively.
% \section{Conclusion}

% In this work, we introduced a novel training-based head pruning method and a simple training-free FFN pruning method for vision-language models, leveraging information bottleneck metrics. In the former method, We developed a loss function that combines eRank for discarding redundant information and KS distance for preserving meaningful information, which leads to an effective pruning. In the latter method, we design a algorithm to calculate a retaining rank by a deviation threshold. our method effectively reduces model size while preserving performance. The proposed approach allows for fine-grained pruning, ensuring efficient model compression with minimal information loss.

% Future work could extend this method to include language module components, enhancing pruning effectiveness. Combining this approach with other compression techniques, such as quantization and knowledge distillation, could further optimize model efficiency. Exploring new metrics aligned with information bottleneck theory for pruning effectiveness may refine our approach and contribute to more robust pruning frameworks. These advancements could lead to highly efficient multimodal models, especially for real-world applications on edge devices with limited computational resources.

\section{Conclusion}

We introduced \texttt{InfoPrune}, an information-theoretic framework for compressing vision-language models through a training-based attention head pruning scheme and a training-free FFN compression approach. 
Grounded in the Information Bottleneck principle, the head pruning method integrates effective rank (eRank) for redundancy reduction and Kolmogorov--Smirnov (KS) distance for information preservation, enabling adaptive and interpretable pruning. 
For FFNs, a low-rank approximation based on a deviation threshold determines the optimal retained rank, achieving lightweight structural simplification without retraining. 
Together, these techniques deliver fine-grained compression with minimal performance loss, enhancing efficiency while maintaining semantic fidelity.

In the future, we plan to extend InfoPrune to the language components of multimodal systems to further enhance compression consistency. 
Combining the proposed approach with complementary techniques such as quantization and knowledge distillation could yield additional gains in efficiency. 
Overall, InfoPrune offers a principled and practical path toward highly efficient multimodal models, enabling their deployment on edge devices with limited resources.
{
    \small
    \bibliographystyle{ieeenat_fullname}
    \bibliography{main}
}

% \input{appendix}
\clearpage
\appendix
\setcounter{page}{1}
\maketitlesupplementary

\section*{Appendix A: Detailed Proof of Redundancy Reduction and eRank Optimization}

This appendix provides the detailed derivation of the mutual information difference and its relation to eRank minimization. The goal is to ensure that the pruned output matrix contains less redundant information, which can be formalized as:

\[
I(X; Z_S) - I(X; Z) < 0
\]

We decompose the difference in mutual information, using the relationship between mutual information and entropy, as shown in Eq.~\ref{entropy subtraction}. The mutual information difference can be expressed as:

\begin{align}
&I(X; Z_S) - I(X; Z) \notag\\
&= H(Z_S) - H(Z_S | X) - \left[ H(Z) - H(Z | X) \right]\label{entropy subtraction}\\
&= H(Z_S) - H(Z) \label{Mutual information subtraction} \\
&= \log \mathrm{eRank}(Z_S) - \log \mathrm{eRank}(Z) \label{erank subtraction}
\end{align}

From the assumption that the attention mechanism outputs \( Z = f(X) \) as a deterministic function of the input \( X \), we have \( H(Z|X) = 0 \) and \( H(Z_S|X) = 0 \). This simplification leads to the expression in Eq.~\ref{Mutual information subtraction}. Additionally, we show that for a covariance matrix of normalized vectors, \( \mathrm{eRank}(\Sigma_S) \) is equivalent to \( \exp(H(\Sigma_S)) \), which is captured in Eq.~\ref{erank subtraction}.

\section*{Appendix B: Detailed Proof of Preserving Meaningful Information}

\subsection*{A Discussion on KS Distance}

Starting from the definition of the empirical distribution function, we now provide a detailed analysis and derivation of the Kolmogorov–Smirnov (KS) distance. Given a sample \( X_1, X_2, \ldots, X_n \), the empirical distribution function \( F_n(x) \) is defined as:
\[
F_n(x) = \frac{1}{n} \sum_{i=1}^n I(X_i \leq x)
\]
where \( I(X_i \leq x) \) is the indicator function, which equals 1 if \( X_i \leq x \) and 0 otherwise.

\subsubsection*{Case 1: \( x \in (\lambda_{k+1}, \lambda_k) \)}
In this case, the top \( k \) singular values \( \lambda_1, \ldots, \lambda_k \) satisfy \( \lambda_i > x \), while the remaining \( n - k \) singular values satisfy \( \lambda_j \leq x \). Therefore:
\[
F_{1,r_1}(x) = \frac{n-k}{n}, \quad F_k(x) = 0
\]
\[
\left| F_{1,r_1}(x) - F_k(x) \right| = \frac{n-k}{n}
\]

\subsubsection*{Case 2: \( x \leq \lambda_1 \) or \( x \geq \lambda_n \)}
When \( x \leq \lambda_1 \), all singular values are greater than \( x \), so:
\[
F_{1,r_1}(x) = 0, \quad F_k(x) = 0
\]
\[
\left| F_{1,r_1}(x) - F_k(x) \right| = 0
\]

When \( x \geq \lambda_n \), all singular values are less than or equal to \( x \), hence:
\[
F_{1,r_1}(x) = 1, \quad F_k(x) = 1
\]
\[
\left| F_{1,r_1}(x) - F_k(x) \right| = 0
\]

\subsubsection*{Case 3: \( x \in (\lambda_n, \lambda_{k+1}) \)}
In this case, \( x \) lies between a non-retained singular value \( \lambda_j \) and another smaller value, but is still smaller than all top-\( k \) retained values:
\[
F_{1,r_1}(x) = \frac{n - m}{n}, \quad F_k(x) = 0
\]
\[
\left| F_{1,r_1}(x) - F_k(x) \right| = \frac{n - m}{n} < \frac{n-k}{n}
\]

\subsubsection*{KS Distance Summary}
Combining all cases, the maximum discrepancy occurs in Case 1. Thus, the KS distance is given by:
\[
\mathcal{D}_{\text{KS}} = \sup_x \left| F_{1,r_1}(x) - F_k(x) \right| = \frac{n-k}{n}
\]

\subsection*{Singular Value Decomposition and Pruning Strategy}

We now return to the singular value decomposition (SVD) of the intermediate representation matrix \( Z \), which can be written as:
\[
Z = U \Sigma_Z V^\top, \quad \Sigma_Z = \text{diag}(\sigma_1, \sigma_2, \ldots, \sigma_{r_1})
\]
where the singular values are ordered such that \( \sigma_1 \geq \sigma_2 \geq \cdots \geq \sigma_{r_1} > 0 \), and \( r_1 = \text{rank}(Z) \). These singular values quantify the contribution of each dimension to the representation. Larger singular values capture more informative components related to the target variable \( Y \), while smaller singular values are often associated with noise or redundancy.

To perform information-preserving pruning, we select an integer \( k < r_1 \) and retain the top \( k \) singular values. This is consistent with the optimal low-rank approximation framework that preserves the most informative subspace:
\[
\Sigma_{Z_S} = \text{diag}(\sigma_1, \sigma_2, \ldots, \sigma_k)
\]

The resulting pruned matrix \( Z_S \) therefore retains the dominant subspace of the original representation \( Z \), maximizing the mutual information \( I(Y; Z_S) \) with respect to the predictive target \( Y \), and minimizing information loss through structured dimensionality reduction.

\section*{Appendix C: Detailed Proof of Adaptive FFN Pruning}

According to the Eckart-Young theorem, we have $\|W - \hat{W}_k\|_F = \sqrt{\sum \limits_{i = k + 1}^r \sigma_i^2}$, where $r$ is the rank of matrix $W$. Hence:
\begin{equation}
	\sum_{i = k + 1}^r \sigma_i^2 \le \epsilon^2 \|W\|_F^2 = \epsilon^2 \sum_{i = 1}^r \sigma_{i}^2.
\end{equation}
Since $\sum \limits_{i = k + 1}^r \sigma_i^2 = \sum \limits_{i = 1}^r \sigma_{i}^2 - \sum \limits_{i =1}^k \sigma_{i}^2$, we get:
\begin{equation}
% \label{FFN-pruning}
	\sum_{i = 1}^k \sigma_i^2 \geq (1 - \epsilon^2) \sum_{i=1}^r \sigma_i^2.
\end{equation}
% Q.E.D.

% \end{document}

\end{document}